\documentclass{elsarticle}
\usepackage{palatino,latexsym,natbib}
\usepackage{lineno,hyperref}
\modulolinenumbers[5]

\journal{Journal of Artificial Intelligence}
\usepackage{pst-eucl}
\psset{opacity=.2}
\usepackage{listings,url}
\usepackage{graphicx,paralist,subcaption}

\newtheorem{definition}{Definition}
\usepackage{booktabs}
\usepackage{color,amssymb,amsmath,algorithm,algorithmicx,algpseudocode}

\usepackage{xargs} 
\usepackage[colorinlistoftodos,prependcaption,textsize=tiny]{todonotes}
\newcommandx{\todoa}[2][1=]{\todo[linecolor=purple,backgroundcolor=purple!25,bordercolor=purple,#1]{#2}}
\newcommandx{\todomw}[2][1=]{\todo[linecolor=olive,backgroundcolor=olive!25,bordercolor=olive,#1]{#2}}
\newcommandx{\todom}[2][1=]{\todo[linecolor=blue,backgroundcolor=blue!25,bordercolor=blue,#1]{#2}}
\usetikzlibrary{arrows, quotes,backgrounds,calc}
\usepackage{tikz}
\usetikzlibrary{arrows.meta}
\usetikzlibrary{arrows,shapes,decorations,automata,backgrounds,petri}
\tikzset{point/.style = {fill=black,circle,inner sep=0.7pt}}
\tikzset{
  graph vertex/.style={
    circle,
    draw,
  },
    graph vertexx/.style={
    draw,
  },
  graph directed edge/.style={
    ->,
    >=stealth,
    thick,
  },
  graph tree edge/.style={
    graph directed edge
  },
  graph forward edge/.style={
    graph directed edge,
    every edge/.style={
      edge node={node [fill=white,font=\scriptsize] {f}},
      loosely dotted,
      draw,
    },
  },
  graph back edge/.style={
    graph directed edge,
    every edge/.style={
      edge node={node [fill=white,font=\scriptsize] {b}},
      densely dotted,
      draw,
    },
  },
  graph cross edge/.style={
    graph directed edge,
    every edge/.style={
      edge node={node [fill=white,font=\scriptsize] {c}},
      dotted,
      draw,
    },
  },
}

\usepackage{changes}
\definechangesauthor[name={R G}, color=red]{rg}
\setremarkmarkup{(#2)}

\begin{document}
\begin{frontmatter}
\title{Is perturbation an effective restart strategy?}  

\author[mymainaddress]{Aldeida Aleti\corref{mycorrespondingauthor}}
\ead{aldeida.aleti@monash.edu}
\cortext[mycorrespondingauthor]{Corresponding author}

\author[mymainaddress]{Mark Wallace}
\ead{mark.wallace@monash.edu}

\author[markus]{Markus Wagner}
\ead{markus.wagner@adelaide.edu.au}

\address[mymainaddress]{Faculty of Information Technology, Monash University, Australia}
\address[markus]{School of Computer Science, The University of Adelaide, Australia}

\begin{abstract}

Premature convergence can be detrimental to the performance of search methods, which is why many search algorithms include restart strategies to deal with it. 
While it is common to perturb the incumbent solution with diversification steps of various sizes with the hope that the search method will find a new basin of attraction leading to a better local optimum, it is usually not clear how big the perturbation step should be.
We introduce a new property of fitness landscapes termed {\em Neighbours with Similar Fitness} and we demonstrate that the effectiveness of a restart strategy depends on this property.

\end{abstract}
\begin{keyword}
Search algorithms, escaping local optima, restart strategies, perturbation.
\end{keyword}
\end{frontmatter}

\section{Introduction}

A wide variety of techniques has been developed for tackling large scale combinatorial optimisation problems. Search methods, such as Genetic Algorithms and Simulated Annealing are typically used to achieve the required scalability in challenging problems for which it is hard to find optimal, or even just ``good enough' solutions. The majority of these methods involve steps where the state of the algorithm is modified in some way to escape a local optimum. The aim is to avoid premature convergence, which is when the search method converges (usually very early in the search) to a local optimum of poor quality~\cite{SCHUURMANS2001121,ZIVAN20141}.

 Previous research has shown that the performance of search strategies is affected by the structure of the fitness landscape~\cite{moser2016investigating, moser2016identifying}. A fitness landscape is defined by three components: i) the search space, which is the set of all candidate solutions, ii) the fitness function, which assigns a fitness value to each solution, and the neighbourhood operator, which defines how solutions are connected, and as a results how the search strategy can traverse the landscape. An example of a neighbourhood operator is the 1-flip operator, which flips the value of a bit in a bitstring representation. In this fitness landscape, all solutions that differ by one bit are neighbours. A different neighbourhood operator creates a different landscape, with different landscape structures, such as the number of local optima and plateaus, affecting how well a particular search method performs. To study this problem, exciting research focuses on fitness landscape characterisation metrics~\cite{moser2016identifying}, which measure properties of the fitness landscape and relate them to the effectiveness of search algorithms.     

In this paper, we argue that the properties of a fitness landscape have an impact on restart strategies used by search methods. Most search algorithms implement restart during the search, as a mechanism for resetting the search and preventing premature convergence~\cite{lourencco2019iterated}, however, there is little understanding on when these methods work. We focus on a notable restart strategy, known as random perturbation, where the local optimum is perturbed by applying diversification steps of various sizes. The size of perturbation is a parameter that has to be tuned, and in this paper we show that its effectiveness depends on the features of a landscape. To this end, we examine the effectiveness of the strength of perturbation, and analyse conditions under which local perturbation does not help escape the local optimum. To enable the analysis, we introduce a new property of fitness landscapes, \textit{Neighbours with Similar Fitness} or NSF, which means that neighbours in the search space tend to have similar fitness compared to non-neighbours. A neighbour is a solution that can be reached in one step by the search method from its current position. An NSF neighbourhood is one that tends to link solutions with similar fitness.

We formalise the definition of NSF, and through an experimental evaluation on 14,000 fitness landscapes with and without this property show that random perturbation is not effective on NSF landscapes, even when the perturbation is large. We argue that NSF generally holds for combinatorial optimisation problems and the search operators designed for algorithms that tackle them, hence we recommend a random restart strategy as a default, which picks at random a new solution in the search space. The dataset and code used in this paper are available online at \url{https://github.com/aaleti/NeighboursSimilarFitness}

\section{Fitness Landscapes and Their Properties}

The suitability of a search method for solving an optimisation problem instance depends on the structure of the fitness landscape of that instance. A fitness landscape in the context of combinatorial optimisation problems refers to 
\begin{compactitem}
\item the search space $S$ of all solutions,
\item an ordered set of fitness values $V$
\item the fitness function $f: S{\rightarrow}V$, which maps solutions to values, and
\item a neighbourhood operator, which assigns to each solution $s{\in}S$ a set of neighbours $N(s){\subseteq}S$.
\end{compactitem}

\begin{figure}
    \centering
 \begin{tikzpicture}[font=\footnotesize]
    \coordinate (0000) at (0,7);
    \coordinate (0001) at (-3,6);
    \coordinate (0010) at (-1,6);
    \coordinate (0100) at (1,6);
    \coordinate (1000) at (3,6);
    \coordinate (0011) at (-3.75,5);
    \coordinate (0101) at (-2.25,5);
    \coordinate (1001) at (-0.75,5);
    \coordinate (0110) at (0.75,5);
    \coordinate (1010) at (2.25,5);
    \coordinate (1100) at (3.75,5);
    \coordinate (0111) at (-3,4);
    \coordinate (1011) at (-1,4);
    \coordinate (1101) at (1,4);
    \coordinate (1110) at (3,4);
    \coordinate (1111) at (0,3);
    \foreach \x in {0000,0001,0010,0100,1000,0011,0110,1100,1010,0101,1001,0111,1101,1011,1110,1111}{
    \node[point] at (\x) {};
    \node[] at ($(\x)+(-0.25,0.2)$) {$\x$};
    } 
    \draw (0000) -- (0001) (0000)--(0010)   (0000)-- (0100) (0000)--(1000);
    \draw (0001)--(0011) (0001)--(0101) (0001)--(1001);
    \draw (0010)--(0011) (0010)--(0110) (0010)--(1010);
    \draw (0100)--(0101) (0100)--(0110) (0100)--(1100);
    \draw (1000)--(1001) (1000)--(1010) (1000)--(1100);
    \draw (0111)--(0011) (0111)--(0101) (0111)--(0110) (0111)--(1111);
    \draw (1011)--(0011) (1011)--(1111) (1011)--(1001) (1011)--(1010);
    \draw (1101)--(0101) (1101)--(1001) (1101)--(1100) (1101)--(1111);
    \draw (1110)--(0110) (1110)--(1010) (1110)--(1100) (1110)--(1111);
  \end{tikzpicture}
    \caption{A hypothetical search space of four Boolean decision variables. The solid lines represents pairs of solutions that are considered neighbours, under the assumption of a 1-flip neighbourhood operator.}
    \label{fig:searshspace}
\end{figure}
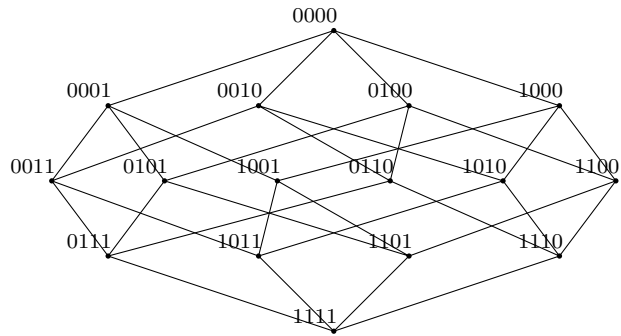

Typically the search space is defined by a set of decision variables, and (without loss of generality) a set of possible values that each decision variable can take.
A binary decision variable, for example, can take just two possible values. An example of a neighbourhood operator is the 1-flip operator, which assigns neighbours to solutions if they differ in the value of just one binary decision variable. This example is illustrated in Figure~\ref{fig:searshspace}. A different neighbourhood operator, such as the swap operator, would result in different solutions being connected, and thus impacting the structure of the fitness landscape.

\begin{figure}[!ht]
\centering
\includegraphics[width=0.3\linewidth]{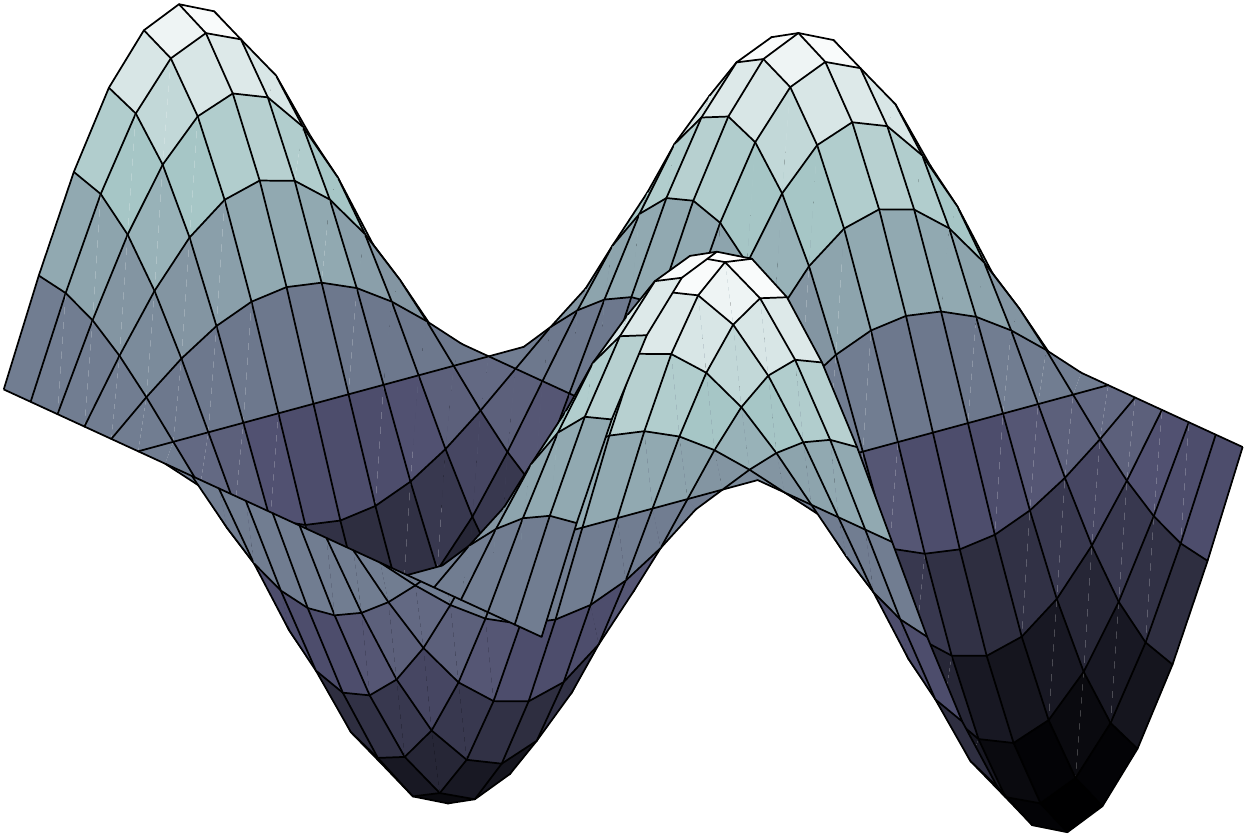}\hspace{10mm}\includegraphics[width=0.43\linewidth]{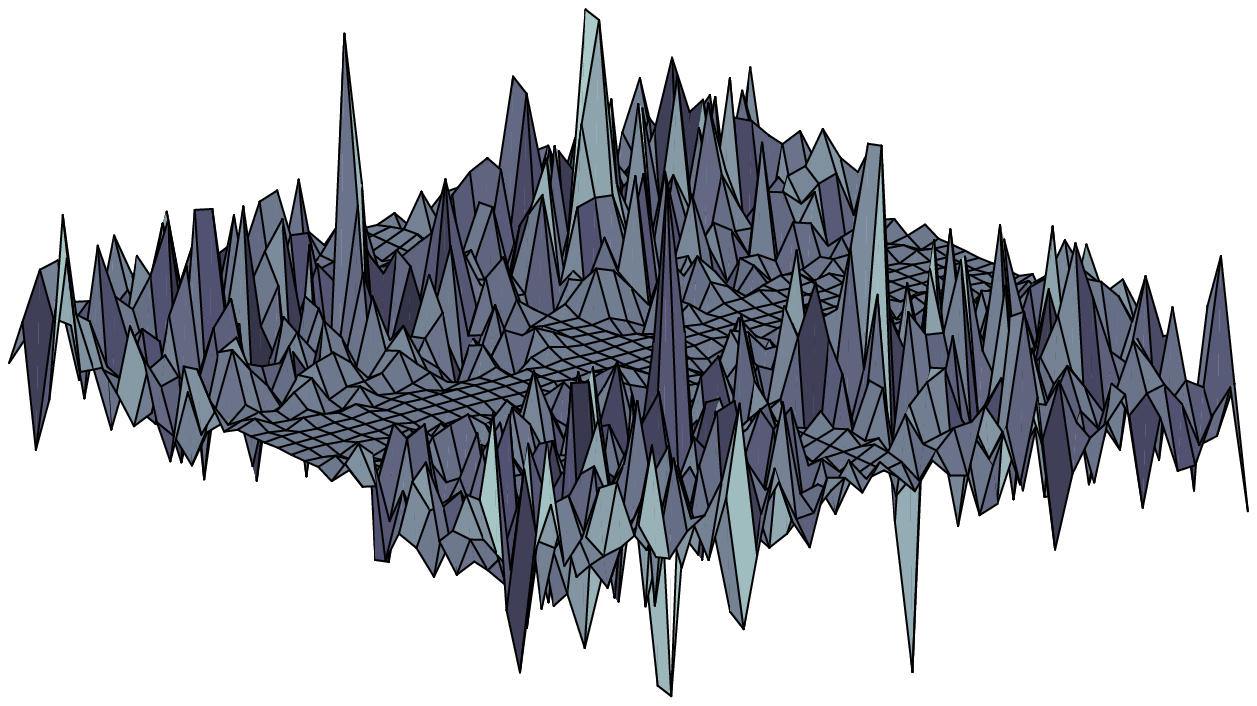}
\caption{An illustration of two fitness landscapes with very different structures. These two landscapes represent the same problem and search space, but are constructed using two different neighbourhood operators.\label{fig:FL}}
\end{figure}

The choice of neighbourhood operator affects how the landscape is structured and its properties. Search methods rely on gradients in the landscape, hence their effectiveness and efficiency is affected by the structure of the landscape. Imagine the two landscapes plotted in Figure~\ref{fig:FL}. A search method such as simulated annealing would have no trouble finding the optimal solution in the smooth landscape on the left, while struggling in the rugged landscapes on the right. 

Fitness landscape characterisation is an area that addresses this problem, by devising different measures for estimating landscape properties and relating them to the effectiveness of search algorithms. Some of these methods are approximate, such as metrics based on random walks and sampling, while others require complete knowledge of the search space, e.g., Local Optima Networks (LONs). In the next subsection, we present an overview of this area. The NSF property presented in this paper is a new property of fitness landscapes and a reliable indicator of whether perturbation is an effective restart strategy or not. 
 
\subsection{Landscape Properties Based on Random Walks}

Random walks can reveal a great deal about a landscape. A \emph{random walk} is a local search (LS) path, where the successors are chosen randomly from the neighbours of their predecessors. Firstly, the values of the solutions along the random walk, and in particular how they vary as the length of the random walk grows (compared with the standard deviation in fitness values across the whole search space)~\cite{Weinberger90,Weinberger91b}, can suggest whether LS is likely to work well on this landscape~\cite{Angel98}. This measure, however, fails to detect neutrality in fitness landscapes, which is a challenging feature for gradient-exploiting algorithms~\cite{moser2016identifying}. 

Secondly the frequency with which the value changes during a random walk by more than a given threshold~\cite{Vassilev00} provides useful insights into fitness landscapes. This measure is an estimate of the diversity of the local optima and the behaviour of the fitness landscape.

Thirdly, using the distance evaluator and some estimate of the distance from a solution to a globally optimal solution~\cite{Jones95a}, the random walk can suggest to what extent solution fitness corresponds with distance to the optimal solution. This measure is claimed to predict the success of genetic algorithms on this landscape~\cite{Vassilev00,Merz00,Czech08,Merz04}, though this is debated~\cite{Altenberg97,Quick98,Naudts00}.
 
\subsection{Landscape Properties Based on Local Optima}

Local optima, their distribution, and the shape of the basins of attraction have been subject to extensive research. Local Optima Networks (LON)~\cite{daolio2010local} and Predictive Diagnostic Optimisation (PDO) are two representative methods. While both methods use local search to map the landscape, they differ in the information they process and how the mapping is done. PDO is a landmarking method (the mapping is performed during the optimisation), while LON is a characterisation method (the main aim is not optimisation, but estimating the structure of the landscape).

\paragraph{Local Optima Network}

Local Optima Network (LON)~\cite{daolio2010local} models the fitness landscape as a graph whose vertices are local optima linked with weighted edges that allow the calculation of the probabilities of reaching one local optimum from another. A steepest descent is used to determine the local optima and therefore define a basin of attraction for each of these optima. The size of the basin is given by the cardinality of the set of all solutions that are part of the basin.

LON is a compact representation of the fitness landscape. 
Using this compact model, different features of the landscape can be measured (inspired from complex systems), such as degree distribution, clustering coefficient, shortest path length, disparity, and community structure.

LON served as an instrument to study many fitness landscapes. Subject of one of the earliest studies were different problem instances~\cite{daolio2010local} of the Quadratic Assignment Problem (QAP). It showed that the search difficulty increases with the number of local optima (vertices in the graph) and the problem dimension. It was also noted that uniform problem instances, which have largely similar basins of attraction, produced considerably larger graphs compared to problem instances which have more rugged fitness landscapes. The scalability of LON to large problems is an issue with its general applicability, which might be solved with an appropriate sampling technique.

\paragraph{Predictive Diagnostic Optimisation (PDO)}

PDO~\cite{moser2016investigating,moser2016identifying} is a landmarking method which characterises the fitness landscape while optimising the problem with a local search. PDO projects the quality of the local optima to be expected from a local search given an arbitrary point in the basin of attraction. 

The algorithm starts with a learning phase, where it optimises randomly selected solutions locally, with the goal of finding all predictors necessary to describe all optima encountered in the problem within a tolerance of error. A predictor is created by applying a local search and has two parts: i) the fitness improvement in the early stages of the search, and ii) the subsequent fitness improvement once a local optimum has been found. The idea is that the initial fitness improvement (i.e., the gradient at the start of the search) should provide an indication of the fitness of the local optimum.  A new predictor is created during the learning phase whenever the existing predictors do not describe a new local optima with the expected accuracy. The discovery phase ends when no new predictors have been created in a predefined number of iterations.

In the testing phase, the method checks whether the pool of predictors can accurately describe the shapes of the basins of attraction in the fitness landscape. Similar to the learning phase, random solutions are created and improved locally, and the improvement in the first step helps in selecting the closest-matching predictor. 

After the local search is complete, the prediction made by the representative predictor is used to determine the prediction error (PE) as the difference between actual and expected fitness. Low prediction error means that the fitness landscape is homogeneous, with regular and well-shaped basins of attraction.

\section{Strategies for Dealing with Premature Convergence}

A wide range of different strategies has been devised in order to prevent search methods from converging early in the search to local optima of poor quality. In the following, we provide a brief overview of strategies involving perturbations that range from small perturbations to complete restarts, and that cover problem-agnostic black-box approaches as well as instance-specific white-box approaches.

\subsection{Classical and partial restarts}

Nowadays, stochastic search algorithms and randomized search heuristics are frequently restarted: If a run does not conclude within a pre-determined limit, we restart the algorithm~\citep{Mar:bc:03,Lou-Mar-Stu:bc:10}. 
This was shown to help avoid heavy-tailed runtime distributions~\cite{GomesSCK00}. Due to the added complexity of designing an appropriate restart strategy for a given target algorithm, the two most common techniques used are to either restarts with a certain probability at the end of each iteration, or to employ a fixed schedule of restarts. 

Some theoretical results exist on how to construct optimal restart strategies. For example, Luby et al.~\cite{luby1993} showed that, for Las Vegas algorithms with known run time distribution, there is an optimal stopping time in order to minimize the expected running time. They also showed that, if the distribution is unknown, there is a universal sequence of running times given by (1,1,2,1,1,2,4,1,1,2,1,1,2,4,8,...), 
which is the optimal restarting strategy up to constant factors. Schuurmans et al.~\cite{SCHUURMANS2001121}, on the other hand, identify a series of measurable characteristics of local search behavior that are predictive of problem solving efficiency in incomplete SAT procedures. Building upon these findings, the authors introduce a new restart strategy for their search method, which they call ``smoothed descent and flood''. While Schuurmans et al.~\cite{SCHUURMANS2001121} measure the behaviour of search, we measure the structure of the problem and focus on how the structure of the fitness landscape impacts the effectiveness of restart strategies. 

Fewer results are known for the optimization case. A gentle introduction to practical approaches for such restart strategies is given by Marti~\cite{Mar:bc:03} and Lourenco et al.~\cite{Lou-Mar-Stu:bc:10}, but few theoretical results are known \cite{jansen2002dynrestartsEA,Sch-Tey-Tey:c:12}. Particularly for the satisfiability problem, several studies make an empirical comparison of a number of restart policies~\citep{Biere08, Huang2007}. These show the substantial impact of the restart policy on the efficiency of SAT solvers. 
In the context of satisfiability problems this might be unsurprising as state-of-the-art SAT and CSP solvers speed up their search by learning ``no-goods'' during backtracking~\cite{CireKS14}.

Related to this is the work of Ansotegui et al.~\cite{Ansotegui2017dialectic} who defined a new state-of-the-art in MaxSAT solvers, based on their success in the 2016 MaxSAT evaluation. They created a reactive search which includes in total 84 parameters, which were tuned using a hyperparameter tuner. Among these parameters are the lower bound $a_l$ and upper bound $a_u$ on the percentage of variables to be changed to construct a starting point for a local search. However, as neither the values nor the effects were presented and isolated, the actual impact of the perturbation is unclear.

\subsection{Mutation as perturbation}

While partial and complete restarts cover a wide range of perturbation strength, single applications of a neighbourhood operator can also be seen as perturbations, albeit very small ones.

Let us consider problems where solutions are represented as bitstrings of length $n$. In randomised search heuristics, it is very common to create new solutions based on an existing one by randomly flipping each bit independently and with probability $1/n$. For theoretical analyses, this has the welcomed property that, in expectation, exactly one bit is flipped. 
This variation operator is studied very comprehensively in the area of randomised search heuristics, see e.g. \cite{Neumann2010eaTheoryBook} for an overview.

Interestingly, while there is a chance of flipping more than 1 bit, the number of bits that are flipped (and thus the perturbation strength) is sharply concentrated. It was not until  recently that a major change was proposed: instead of considering the established bitwise $1/n$-mutation (which means that the number of flips bits is drawn from a sharply-concentrated binomial distribution), it was proposed to draw the number of bits flipped from heavy-tailed distributions~\cite{Doerr2017fastGAs}. This has proven especially useful for multi-modal problems, and a range of heavy-tailed distributions of vastly different shapes have been proposed since then theoretically and experimentally~\cite{Friedrich2018heavytailedPPSN,Friedrich2018heavytailedGECCO}. 
Hence, such heavy-tailed mutations can be considered to be partial restarts as well, even though they forego the check for convergence to a local optimum.

Very much related to this is the idea of iterated local search. These approaches iteratively build a sequence of solutions generated by the embedded heuristic, ideally leading to far better solutions than if only repeated random trials of that heuristic are used.
General frameworks for iterated local search can accommodate partial restarts for optimization problems. 
For example, Louren{\c{c}}o et al.~\cite{lourencco2019iterated} acknowledge that ``a good perturbation transforms one excellent solution into an excellent starting point for a local search'', however, the authors only provide anecdotal evidence as well as a few results from experimental investigations -- also, no connection to a property of the landscape is established. 

In this present article, we address this gap between theory and practise by introducing a landscape property in the next section.

\section{Neighbours with Similar Fitness} \label{sec:con}

Neighbours with Similar Fitness (NSF) is a property of fitness landscapes which states that two neighbouring solutions have a higher probability of having similar fitness values than two non-neighbouring solutions. We represent a combinatorial optimisation problem $(S,F)$ as a finite search space $S$ with a finite range of fitness values $\{F(s): s \in S\} \subseteq \{v_{min} \ldots v_{max}\}$. 
Without loss of generality, we consider maximisation problems in the following, i.e., solutions with higher fitness values are considered to be better. 

We define $|F_v|$ to be the number of solutions in the search space with fitness value $v$. 
The search space size is equal to
\begin{equation}|S| = \sum\limits_{v \in V} |F_v|.
\end{equation}
Given a solution with fitness $v$ we would like to know the number of solutions with fitness difference by $\delta$, defined as

\begin{equation}
|F_{v,\delta}| = |\{s \in S : f(s) = v \pm \delta\}|
\end{equation}
where $\delta \in \Delta$, and $\Delta$ is the set of all differences among fitness values in $V$.

This tells us, for a solution with a particular fitness, how different the other solutions are. This also allows us to calculate the proportion of solutions that differ in fitness by $\delta$ from a given fitness $v$
\begin{equation}
p_{v,\delta} = \frac{|F_{v,\delta}|}{|S|}
\end{equation}
A fitness landscape associated with the problem $(S,f)$ is defined as $(S,f,N)$, where $N$ is the neighbourhood function which identifies the solutions that can be reached in one search step from each other solution. We refer to the set of neighbours $N_v$ of solutions with fitness $v$:

\begin{equation}
N_v = \{s' \in N(s): s \in S \cap f(s)=v\}.
\end{equation}
Note that this is the union of all the neighbourhoods of solutions with fitness $v$. We define the number of neighbours of solutions with fitness $v$, with fitness greater or lower by $\delta$ to be:
\begin{equation}
|N_{v,\delta}| = |\{s \in N_v : f(s) = v \pm \delta\}|
\end{equation}
The proportion of neighbours (solutions in $N_v$) that differ from $v$ by $\delta$, is then
\begin{equation}
pn_{v,\delta} = \frac{|N_{v,\delta}|}{|N_v|}
\end{equation}

We use the expressions $p_{v,\delta}$ and $pn_{v,\delta}$ in the following definition of the Neighbours with Similar Fitness (NSF) property. The property states that neighbours tend to have similar fitness than non-neighbours. 

\begin{definition}\label{nsf}
A landscape has the \emph{Neighbours with Similar Fitness} (NSF) property if for all fitness values $v$ the value of the expression 
$ pn_{v,\delta} - p_{v,\delta}$
decreases monotonically with increasing $\delta \in \Delta$, and
$\sum_{\delta \in \Delta} pn_{v,\delta} - p_{v,\delta} \geq 0$
\end{definition}

This property means that for each fitness value $v$
\begin{compactitem}
\item
the probability that a neighbour of $s$ has fitness $f(s)$ close to $v$ is higher than the probability of a solution in the search space as a whole having fitness close to $v$,
\item
as the fitness value difference $\delta \in \Delta$ grows, the difference between the probability $pn_{v,\delta}$ of a neighbour with fitness differing by $\delta$ from $v$, and the probability $p_{v,\delta}$ of a solution in the whole search space with fitness differing by $\delta$ from $v$, monotonically decreases.
\end{compactitem}

We argue that the fitness landscapes of many real-world problems satisfy the NSF property. As an example, let us consider the TSP problem, whose fitness function is a sum of distances. The number of terms in the sum is the number of cities in the TSP. 
The so-called 2-swap neighbourhood for symmetric TSPs changes only two distances in the sum -- no matter how many cities there are in the TSP. This is, in fact, the smallest change possible (in terms of the number of distances changed) while maintaining the constraint that the tour must be a cycle.  
By ensuring that $N-2$ distances remain the same (where $N$ is the number of cities), the 2-swap generates neighbours of which many can be considered to have similar fitness.

For many other problems, the neighbourhood operator is defined so as to change only a few terms in the sum which expresses the fitness function.  
For example, the maximum satisfiability problem is to find an assignment to truth variables that minimises the number of unsatisfied clauses. A clause is just a disjunction of some truth variables (or none) and some negated truth variables (or none). One way of finding a neighbour for this problem is by changing the truth value of one variable ("flipping" a variable).
If the clause length is restricted to just three (variables and negated variables), and if there are, say, 100 truth variables in the problem, then only 
$1-\frac{99^3}{100^3} = 0.03$ of the clauses are likely to contain any given variable.
Thus after flipping a variable 97\% of the clauses will remain unchanged.
Consequently the fitness of a neighbour found by flipping a variable is likely to be similar to the original fitness.

The same argument applies to most problems whose fitness function is a sum of terms in which the number of terms increases with the number of variables in the problem.
Any neighbourhood operator that changes the value of a single variable, or a small set of variables, will only change the value of a small fraction of the terms in the sum. Consequently neighbours are likely to have similar fitness.

\section{The Probability of Converging to the Global Optimum}

Our hypothesis is that the NSF property has an impact on the effectiveness of different restart techniques, where effectiveness is determined as the ability to escape a local optimum. Our research is focused on stochastic search methods, which traverse the search space by transitioning between neighbouring solutions with a certain probability. To model the behaviour of a search algorithm on a fitness landscape, we use a Discrete Time Markov Chain~\cite{trivedi1982probability}. 

\begin{definition}
A Discrete Time Markov Chain (DTMC) is a discrete-time stochastic process, that can be defined as a set of states and a probability matrix $P$ that represent the probability of transitioning from one state to another. 
\end{definition}

Naturally, the states in the DTMC represent solutions in the fitness landscape. The possible transitions from a solution are to those neighbours of the solution that have a better or the same fitness value. Solutions whose fitness value is locally or globally optimal, have no possible outwards transitions (they are {\em absorbing} states). To illustrate how the DTMC models the behaviour of a search algorithm, consider the following search space:

\begin{table}[!ht]
\normalsize
\centering
\begin{tabular}{lcccccc}
  $s_i$&   $s_1$&$s_2$  &$s_3$  &$s_4$&$s_5$&$s_6$\\
  $f(s_i)$&1&2&3&4 &5 & 6\\
\end{tabular}
\end{table}

where $f(s_i)$ is the fitness of solution $s_i$. We denote the probability that a solution $s_i$ is reached by a search algorithm in one step by changing a solution $s_j$ as $pt(s_i,s_j)$ (probability of transitioning). Next, we construct the \textit{transition matrix} $P=[pt(s_i,s_j)]$ that denotes the transition probabilities for all pairs of solutions $(s_i, s_j) \in S$. Let us assume that the transition matrix for our toy problem is as follows:
\[
P=
\begin{pmatrix}
    0   & 0.25    & 0.25  & 0.25  & 0.25   & 0 \\
    0   & 0      & 0.25 & 0.25  & 0.25   & 0.25 \\
    0   & 0      &0       &0.\overline{3} & 0.\overline{3}   & 0.\overline{3} \\
    0   &0    &0     &0 &0.5   & 0.5 \\
    0      &0       &0       &0    &0   & 1 \\
    1      &0       &0       &0    &0     & 0 \\
\end{pmatrix}
\]

In this example, the probabilities of making a transition to a better or equal neighbour are equal and do not depend on the fitness of the neighbours. Some search methods, however, use a fitness proportionate transition probability, where each neighbour can be reached with a probability proportionate to its fitness. For simplicity, we are using the former strategy to illustrate our approach, however, our approach for calculating the probability of escaping a local optimum can be used with other transition strategies as well.

\begin{figure}[!ht]
\centering
\begin{tikzpicture}[]
\foreach \letter [count=\c] in {1,2,3,4,5,6} {
    \node[graph vertex] (\letter) at ({-61*\c+157.5}:3cm) {$s_{\letter}$};
  }
  
  \begin{scope}[>={Stealth[black]},
              every node/.style={fill=white,circle},
              every edge/.style={draw=black}, font=\scriptsize]]
\path [->] 

(1) edge node {0.25} (2)
    edge[bend left=22] node {0.25} (4)
    edge node {0.25} (5)
(2) edge node {0.25} (3)
    edge node {0.25} (4)
     edge node {0.25} (5)
     edge node {0.25} (6)
  (1) edge node {0.25} (3)
 (3) edge node {0.$\overline{3}$} (4)
     edge node {0.$\overline{3}$} (5)
     edge [bend left=22] node {0.$\overline{3}$} (6)
 (4) edge node {0.5} (5)
     edge [bend left=10] node {0.5} (6)
 (5) edge node {1} (6)
 (6)  edge [loop above] node {1} (6);
 \end{scope}
\end{tikzpicture}
\caption{Transition graph annotated with transition probabilities.}
\label{fig:TG}
\end{figure}
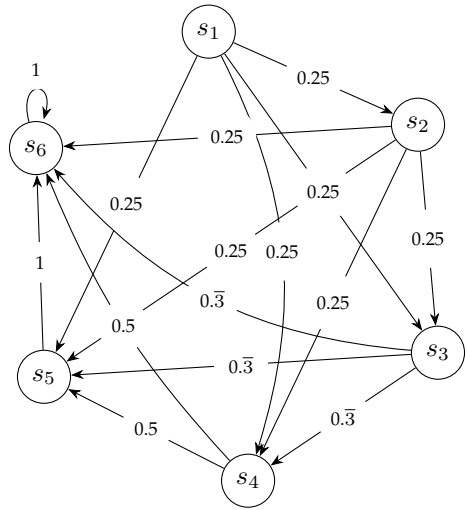

The corresponding transition graph is visualised in Figure~\ref{fig:TG}. This is the DTMC that represents the behaviour of the search algorithm over our fitness landscape.

The DTMC of our example shown in Figure~\ref{fig:TG} has an absorbing state $s_6$, which has no outgoing probabilities, and as a result, the chain can never leave this solution once entered. This is a local or global optimum. Non-absorbing states are called transient states, and a run of a search method corresponds to a random walk in this transition graph. The transition matrix P for an absorbing Markov Chain can be written in \textit{canonical form} as follows~\cite{trivedi1982probability}:

\[
 P=
 \begin{pmatrix}
 Q&R\\
0&I
\end{pmatrix}
 \]

where 
\begin{itemize}
    \item $Q$ is a matrix describing the transition probabilities between transient states, where $pq(s_i,s_j)$ is the probability of transitioning from state $s_i$ to state $s_j$. In other words, this matrix describes the behaviour of the search algorithm before it converges to a local optimum, where each probability represents the likelihood that the search method transitions to solution $s_j$ once it has reached solution $s_i$.
    \item $R$ contains the transition probabilities from transient to absorbing states $pr(s_i,s_j)$, 
    \item and $I$ is the identity matrix.
\end{itemize}

Representing the DTMC transition matrix in a canonical form enables us to estimate the probability of reaching any of the solutions in the search space during the run of a search method. as explained and illustrated below. In our example, $Q, R$, and $I$ are as follows:

\begin{figure}[htbp]
\begin{minipage}[b]{0.4\linewidth}
 \[
 Q=
\begin{pmatrix}
   0   & 0.4    & 0.3  & 0.2  & 0.1 \\
    0   & 0      & 0.4 & 0.3  & 0.2  \\
    0   & 0      &0       &0.5 & 0.3   \\
  0   &0    &0     &0 &0.6    \\
   0      &0       &0       &0    &0   
 \end{pmatrix}\]
 \end{minipage}
\hspace{4mm}
 \begin{minipage}[b]{0.13\linewidth}
\[
R=
 \begin{pmatrix}
   0 \\
 0.1 \\
 0.2 \\
  0.4 \\
  1 
\end{pmatrix}\]
\end{minipage}
 \begin{minipage}[b]{0.4\linewidth}
 \[I=
\begin{pmatrix}
     1   & 0    & 0  & 0  & 0 \\
    0   & 1    & 0  & 0  & 0 \\
     0   & 0    & 1  & 0  & 0 \\
    0   & 0    & 0  & 1  & 0 \\
    0   & 0    & 0  & 0  & 1
 \end{pmatrix}\]
 \end{minipage}
 \end{figure}

Next, we calculate the probability that a solution in a search space is reached (or visited) by the search algorithm in any number of steps, starting from any solution in the search space. Eventually, this will enable us to calculate the probability of reaching the local or global optimum by the search method. The probability $pq^k(s_i,s_j)$ represents the probability of reaching solution $s_j$ from solution $s_i$ in k search steps (transitions). As a results, the matrix $Q^k$ represents the probabilities of reaching transient states in $k$ steps. We can calculate $Q^k$ for any $k$, and the sum of all $Q^k$ gives us the \textit{fundamental matrix} for an absorbing Markov chain as~\cite{sinclair1993markov}:

 \begin{equation}
 N = I + Q + Q^2+Q^3+ \ldots = \sum\limits_{k=0}^{\infty}Q^k = (I-Q)^{-1}
 \end{equation}

The probabilities $pn(s_i,s_j)$ in the matrix $N$ represent the probability that a search method visits a transient state $s_j$ -- i.e., a solution that is not a local or global optimum -- starting from $s_i$ before it is absorbed, i.e.,  before it converges to the optimum. In our toy example, the ``fundamental''  matrix is as follows:

\[
N=
\begin{pmatrix}
1	&	0.25	&	0.3125	&	0.41\overline{6}	&	0.625	\\
0	&	1	&	0.25	&	0.\overline{3}	&	0.5	\\
0	&	0	&	1	&	0.\overline{3}	&	0.5	\\
0	&	0	&	0	&	1	&	0.5	\\
0	&	0	&	0	&	0	&	1	\\
\end{pmatrix}
\]

In our toy example, the matrix $R$ is equal to
\[
R=
\begin{pmatrix}
0\\
0.25\\
0.\overline{3}\\
0.5\\
1
\end{pmatrix}\]

As a reminder, $R$ contains the transition probabilities from transient to absorbing states.
Using the fundamental matrix $N$, we can compute the probability that an absorbing state (global or local optimum) is reached given a starting state, which is equal to:

\begin{equation}
B=NR
\end{equation}
 
In our example, the probability that the search method converges to the global optimum $s_6$ starting from any solution is equal to:

\[
B=
\begin{pmatrix}
1	&	0.25	&	0.3125	&	0.41\overline{6}	&	0.625	\\
0	&	1	&	0.25	&	0.\overline{3}	&	0.5	\\
0	&	0	&	1	&	0.\overline{3}	&	0.5	\\
0	&	0	&	0	&	1	&	0.5	\\
0	&	0	&	0	&	0	&	1	\\
\end{pmatrix} \begin{pmatrix}
0\\
0.25\\
0.\overline{3}\\
0.5\\
1
\end{pmatrix} =
\begin{pmatrix}
1\\
1\\
1\\
1\\
1
\end{pmatrix}
\]

As the matrix $B$ shows, there is a 100\% probability that the global optimum is reached from any solution, that is the search algorithm always converges to $s_6$. This indicates that the fitness landscapes has no local optima, and the global optimum can be reached from any solution.

\section{The Probability of Escaping a Local Optimum}

Restarting the algorithm is a commonly used methods for improving the performance of search methods, such as local search and genetic algorithms~\cite{weise2019improved}. This addresses the entrapment problem, where the search algorithm prematurely converges to local optima. If there is no improvement in the fitness function for a number of iterations, restarting the method with a new position in the search space increases its likelihood to find another local optimum with better quality. 

On the one hand, the perturbation must be big enough to yield a solution which is in the basin of attraction of another local optimum that has not been found previously. On the other hand, if the perturbation is too big, it may not yield a new solution that is any better than what would be found by random picking. The strength of the perturbation can be a parameter of the search strategy, or can be adapted during the search based on features of the landscape.

We conduct a set of experiments to check how the NSF property affects the size of perturbation needed to escape a local optimum. The search space is defined by a bit-string of $M$ binary decision variables $b_1, b_2, ..., b_M$ and a state (i.e. a solution) is defined by setting each variable $b_i$ to 1 or 0. 

Two solutions are neighbours iff they differ in the value of a single variable. For example, for $M=4$, solutions $1010$ and $1000$ are neighbours. 
Accordingly, a move to a neighbour is possible by flipping one variable. A perturbation is a sequence of flips. Thus any given perturbation can be specified as the set of variables that are flipped. A smaller perturbation flips fewer variables. A good perturbation yields a new solution in the basin of attraction of another optimum, resulting in a successful escape from the current local optimum. However, to establish that the new state is in the basin of attraction of a better optimum, this other optimum has to be found. 
A {\em good perturbation} is a set of variables which can be flipped to yield a better solution.

In practise a search algorithm can only seek a good perturbation by changing variables, initially blindly. We therefore model the perturbation problem by preselecting a set of $N$ variables that can be flipped. The subsequent search to reach a better local optimum is a hill-climbing algorithm. In our experiments we model this by selecting the best subset of the chosen variables to flip. 
In summary, our measure of a good perturbation is, given a set of variables that can be flipped, the probability that a better solution can be reached by flipping a subset of these variables.

\subsection{Complete and Restricted Optimisation Problems}
Out of $M$ variables, we choose a subset $b_i,...,b_N$ of $N$ decision variables which are allowed to change, and keep the other variables  fixed. The idea is that the variables in a given set are those a search algorithm can change to escape from a local optimum. 
We call the variables that can change the \textit{chosen set}, and the fixed variables the \textit{fixed set}. 

Any given fitness landscape, over the $2^M$ possible states of the $M$ variables (i.e. solutions), defines a {\bf complete} optimisation problem. Given the fixed values of $N$ the fixed variables, the same fitness landscape defines a {\bf restricted} optimisation problem over the chosen set of $N$ variables.

\begin{figure}[!ht]
\centering
\begin{subfigure}{.45\textwidth}
\centering
 \begin{tikzpicture}[font=\footnotesize]
    \coordinate (000) at (0,0);
    \coordinate (001) at (2,0);
    \coordinate (010) at (0,2);
    \coordinate (011) at (2,2);
    \coordinate (100) at (0.65,0.45);
    \coordinate (101) at (2.65,0.45);
    \coordinate (110) at (0.65,2.45);
    \coordinate (111) at (2.65,2.45);
    \foreach \x in {000,001,010,011,100,101,110,111}{
    \node[point] at (\x) {};
    \node[] at ($(\x)+(-0.25,0.2)$) {$\x$};
    } 
    \draw (000) -- (001) -- (011) -- (010) --cycle;
    \draw (111) -- (110) -- (100) edge[solid] (101) ;
    \draw[solid] (000) -- (100) (101)-- (111) edge[solid] (011);
    \draw (001) -- (101)  (010) -- (110) ;
  \end{tikzpicture}
  \caption{Complete problem, where all the solutions can be visited by the search algorithm. The solid lines represent which solutions are neighbours.}
  \end{subfigure}
  \hspace{9mm}
  \begin{subfigure}{.45\textwidth}
  \centering
   \begin{tikzpicture}[font=\footnotesize]
      \coordinate (000) at (0,0);
    \coordinate (001) at (2,0);
    \coordinate (010) at (0,2);
    \coordinate (011) at (2,2);
    \coordinate (100) at (0.65,0.45);
    \coordinate (101) at (2.65,0.45);
    \coordinate (110) at (0.65,2.45);
    \coordinate (111) at (2.65,2.45);
    \foreach \x in {000,001,010,011,100,101,110,111}{
    \node[point] at (\x) {};
    \node[] at ($(\x)+(-0.25,0.2)$) {$\x$};
    } 
    \draw (000) -- (001) -- (011) -- (010) --cycle;
    \draw [densely dashed](111) -- (110) -- (100) --(101) ;
    \draw[densely dashed] (000) -- (100) (101)-- (111) -- (011);
    \draw[densely dashed] (001) -- (101)  (010) -- (110) ;
  \end{tikzpicture}
  \caption{Restricted problem of $N=2$, where the first variable is fixed to 0. The dashed lines represent solutions that cannot be visited by the search.}
  \end{subfigure}
\caption{Complete and restricted problems. The neighbourhood operator is the 1-flip operator, which connects as neighbours solutions that change in a single bit. The solid lines represent solutions that are neighbours and can be visited by the search algorithm. The dashed lines represent solution pairs that are neighbours in the complete problem, but can not be reached by the search in the restricted problem.}
\label{fig:xxxx}
\end{figure}
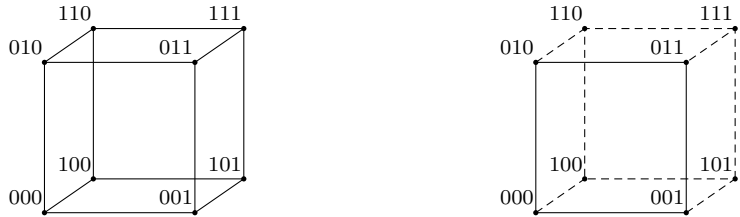

An example of a complete and restricted problem is shown in Figure~\ref{fig:xxxx}. The search space of the complete problem contains all possible combinations of the three variables in the bitstring, i.e., $\{000,001,010,011,100,101,110,111\}$. The restricted problem of $N=2$ has a smaller search space $\{000,001,010,011\}$, composed by solutions that have the first Boolean variable set to 0 (i.e., the fixed variable is equal to 0).

An escape from a local optimum of the complete problem by flipping chosen variables is impossible if the restricted problem is at a global optimum. Therefore, given a fitness landscape, our experimental setup calculates the probability that the restricted problem is at a global optimum. The greater the probability that the restricted problem is at a local, and not a global optimum, the more likely a perturbation restricted to these variables is to successfully escape from the local optimum of the complete problem.

\begin{algorithm}[!ht]
\caption{The generation of NSF and NoNSF fitness landscapes.}\label{alsim}
\begin{algorithmic}[1]
\Procedure{Simulation}{$M$}
\State $\textit{\bf{input}: } N$\Comment{$N$ is the number of chosen variables}
\State $S\gets $ \Call{GenerateBitstrings}{$M$} \label{permutations} \Comment{$2^N$ possible solutions.}
\State $S'\gets$ \Call{RandomReorder}{S}\label{reorder}\Comment{Randomly reorder the solutions. The variables are placed in a list in a random order.}
\State $F, F^* \gets$ \Call{CreateFitnessLandscape}{$S'$, type=\{NSF, NoNSF\}}  \Comment{Randomly generate a fitness function over $S'$, recording the best fitness value generated as $F^*$. For NSF landscapes, the fitness assignment should satisfy the NSF property.}
\EndProcedure
\Procedure{CreateLandscape}{$S$, type=\{NSF, NoNSF\}}
\State $F\gets$\Call{AssignFitness}{S}\Comment{Assign a fitness value to each solution.}
\If{type=NSF}
\State $F\gets$\Call{NSF}{F} \Comment{Enforce NSF property.}
\EndIf
\State $F^*=0$ \Comment{Records the best fitness.}
\For{$r \gets 0$, $r< 1,000, r{++}$}
\State $s \gets $ \Call{RandomPicking}{$S$} \Comment{Randomly choose an initial solution}
\If{\Call{HillClimb}{$s$}$>F^*$}  
\State $F^*=f(s)$\Comment{this optimum is global (among the chosen variables)}
\Else
\State \Return{}\Comment{otherwise this optimum is local (i.e. there is a way to improve the current solution by changing only variables in this set).}
\EndIf
\EndFor
\State \Call \Return{$F^*$}\Comment{Return the best fitness value.}
\EndProcedure
\Procedure{HillClimb}{$s$}
\While{TRUE}\Comment{While the incumbent solution is not the local optimum}
\If{$s_j \in N(s): f(s_j)>f(s)$} 
\State $s \gets s_j$\Comment{Amongst the neighbours with better values choose one.}
\ElsIf{$s_j \in N(s): f(s_j)==f(s)$}
\State $s \gets s_j$ \Comment{If there is no better neighbour, randomly choose a neighbour with the same fitness, that has not previously been encountered in the current hill-climb. Move to the chosen neighbour.}
\Else
\State \Call{Return}{$f(s)$}
\EndIf
\EndWhile
\EndProcedure
\end{algorithmic}
\end{algorithm}

The experiments simulate this situation by modelling the restricted problem as described in Algorithm~\ref{alsim}. Each experiment creates a fitness landscape for the solutions of the restricted problem. These solutions represent the solutions of the complete problem, given the fixed values of the fixed variables.

The experimental software then calculates, for each solution, the probability that a hill climb starting at that solution will reach a global optimum for the restricted problem.
Assuming each initial solution is equally likely, this yields a probability that a local optimum for the complete problem is a global optimum for the restricted problem.

\subsection{NSF and NoNSF Fitness Landscapes}
The experiment randomly creates two kinds of fitness landscapes. The first kind of fitness landscape has a fixed finite set of possible fitness values and one such value is picked without any constraint, at random, for each solution. Such instances are the base case against which a version of NSF will be compared.

The second kind of fitness landscape satisfies a constraint as close as practicable to the NSF property.
It is unclear how one could randomly select members from the set of all fitness landscapes that satisfy exactly the NSF constraint. The constraint used in the experiments requires that the difference in fitness between two neighbouring solutions is never greater than one. This is indeed a much tighter constraint than NSF requires, and as a results satisfies NSF.

Given a problem size $N$ and property \textit{NSF} or \textit{NoNSF}, the process of creating the fitness functions (also shown in Algorithm~\ref{alsim}) works as follows:
\begin{itemize}
\item Create a fitness landscape of $2^N$ bitstrings (solutions), where $N \in [3,9]$ is the number of variables (Step~\ref{permutations} in Algorithm~\ref{alsim}). Two solutions are neighbours if their bitstrings differ in one digit. 
\item Randomly reorder the solutions (Step~\ref{reorder} in Algorithm~\ref{alsim}). The fitness value of each solution is randomly selected by a built-in function that takes as input the variable and the set of values that it can take. Fitness values are assigned at random. NSF landscapes have an additional constraint which states that the fitness of each pair of neighbours differs by at most one. 
\item As soon as the variable has been assigned a specific value, the set of values for the remaining variables in the list are ``pruned'', removing all incompatible values. Thus, under the NSF constraints, if the first variable is given the value 5 (for example), its neighbours will have their sets of values pruned to \{4,5,6\}, and the neighbours of these neighbours will be pruned to \{3,4,5,6,7\}, etc.  
When the second variable is assigned a value, the remaining variables will be pruned further, and again when the third variable is assigned a value, and so on. 
\item If the set of feasible values for a variable becomes empty, then the last assignment is undone  and the variable assigned a different feasible value. However, the NSF constraints are simple enough that this can never happen (because for this class of constraints, propagation enforces global consistency \cite{jeavons1995tractable}).
\item The best fitness value (global optimum) is recorded.
\end{itemize}

For each problem size and property NSF or NoNSF, we create 1,000 distinct landscapes, which in total makes $7\times 2\times 1,000 = 14,000$ fitness landscapes. 

\section{Analysis of Results}

For each reduced fitness landscape, we calculate the probability that absorbing states are reached (i.e., matrix $B$), or in other words, the probability that the search method converges to the global optimum of the restricted problem. This means that, for a particular N, we allow perturbations of size N, and calculate the probability that the search finds the global optimum of the restricted problem. Lower probabilities imply that there are search paths that lead to other local optima of the complete problem with potentially better fitness, which indicate a successful escape of the search. The higher the probability, the harder it is for local search to escape.

\begin{figure*}[!ht]
\begin{subfigure}{.47\linewidth}\includegraphics[width=\linewidth]{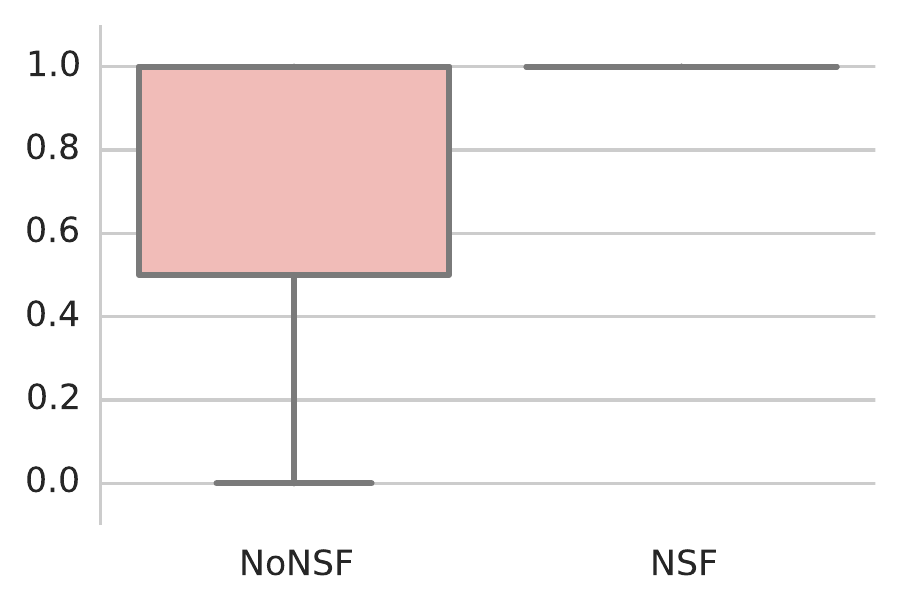}\caption{N=3}\end{subfigure}
\begin{subfigure}{.47\linewidth}\includegraphics[width=\linewidth]{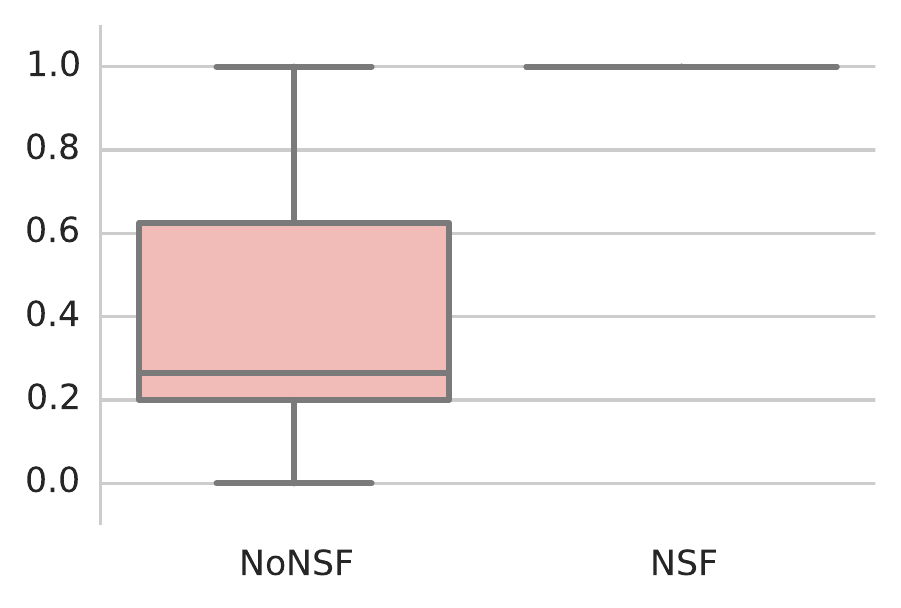}\caption{N=4}\end{subfigure}\\
\begin{subfigure}{.47\linewidth}{\includegraphics[width=\linewidth]{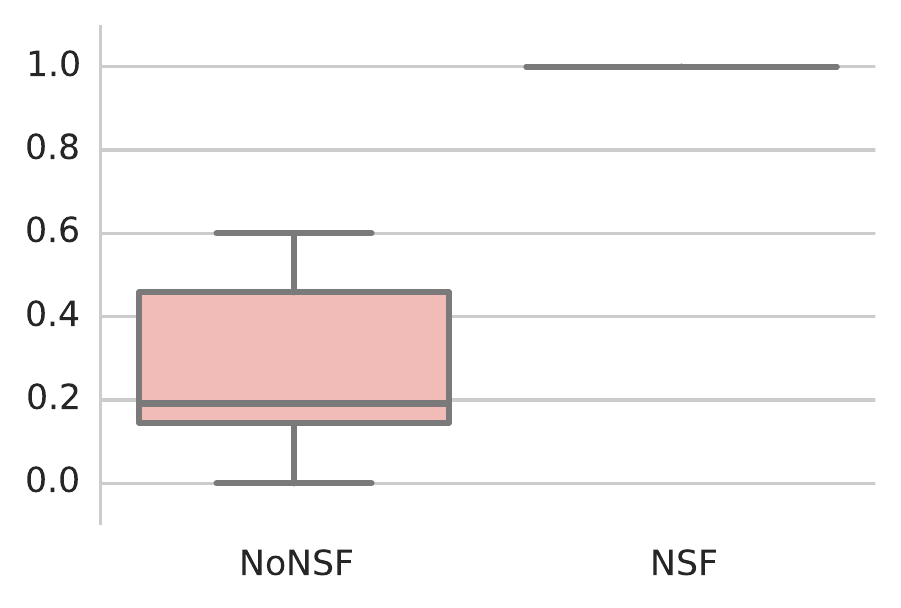}}\caption{N=5}\end{subfigure}
\begin{subfigure}{.47\linewidth}{\includegraphics[width=\linewidth]{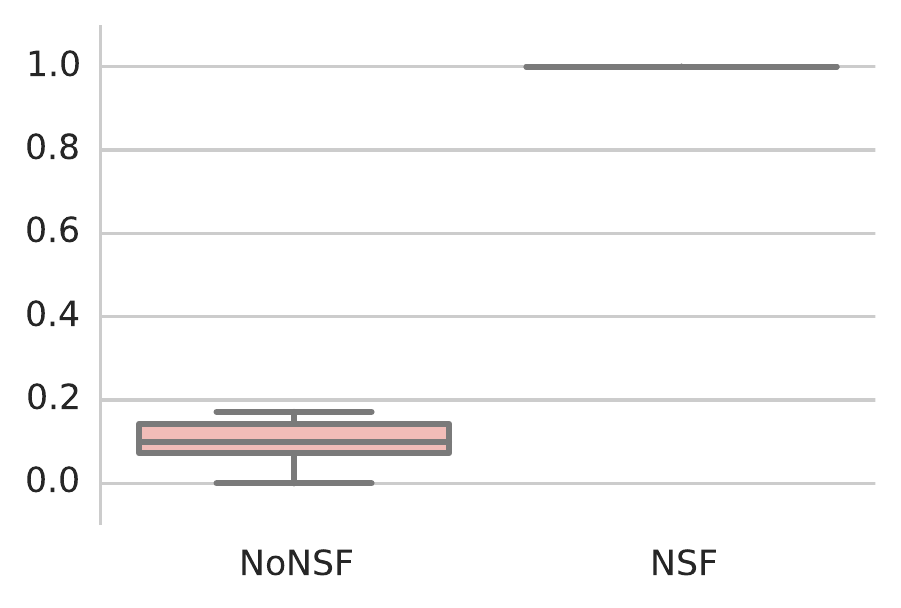}}\caption{N=6}\end{subfigure}\\
\begin{subfigure}{.47\linewidth}{\includegraphics[width=\linewidth]{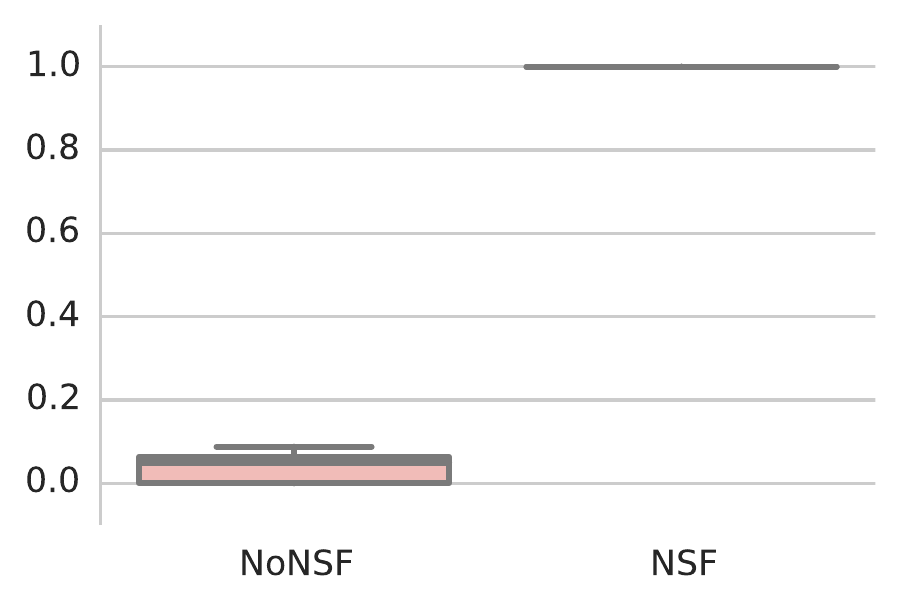}}\caption{N=7}\end{subfigure}
\begin{subfigure}{.47\linewidth}{\includegraphics[width=\linewidth]{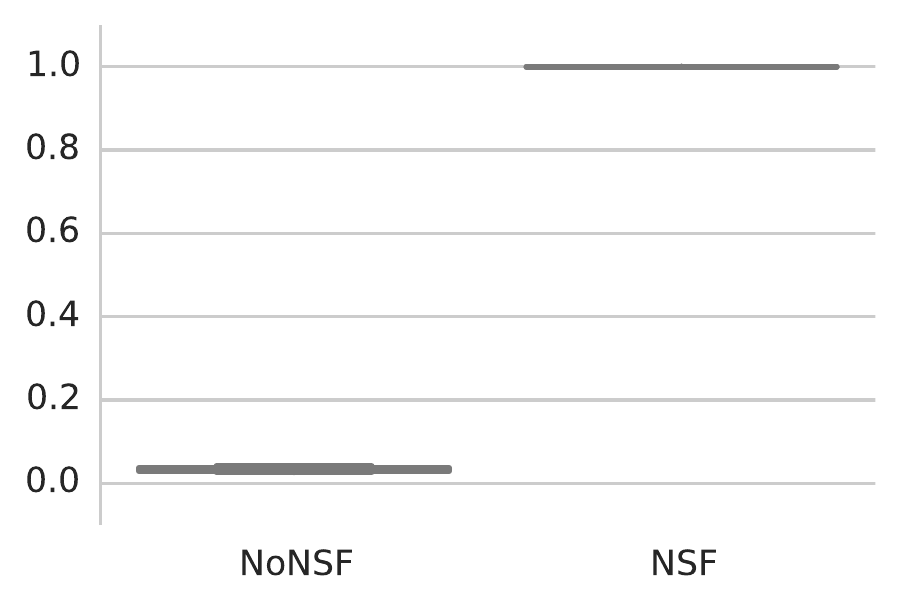}}\caption{N=8}\end{subfigure}\\
\begin{subfigure}{.47\linewidth}{\includegraphics[width=\linewidth]{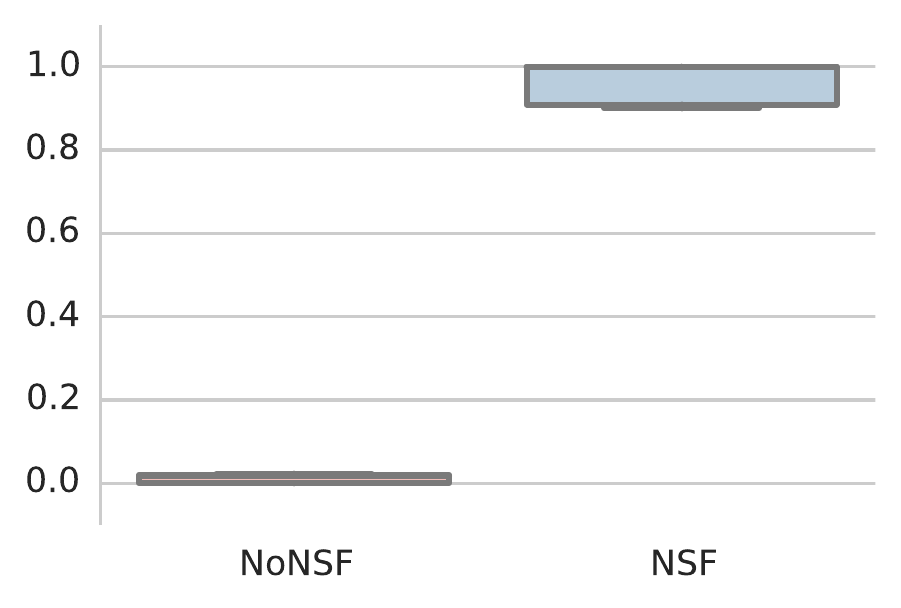}}\caption{N=9}\end{subfigure}
\hspace{3mm}
\begin{subfigure}{.46\linewidth}
\centering
\renewcommand{\arraystretch}{1.1}
\small
  \begin{tabular}{p{1cm}p{1cm}p{1cm}}
  \toprule
 N& NoNSF&NSF\\ \midrule
  3 & 1.00&1.00 \\
  4 & 0.26&1.00 \\
  5 & 0.19&1.00 \\
  6 & 0.10&1.00 \\ 
  7 & 0.05&1.00\\
  8 & 0.04&1.00\\
  9 & 0.02&0.99\\
  \bottomrule
  \end{tabular}
  \caption{Median probabilities.\label{tab:medP}}
\end{subfigure}
\caption{The probability of reaching the global optimum of the restricted problem. \label{fig:boxplots}}
\end{figure*}

Not all absorbing states may be global optima of the restricted problem, hence using matrix B, we estimate the probability of reaching a global optimum by checking that the fitness of an absorbing solution is indeed the maximum fitness for the restricted problem. If multiple solutions with maximum fitness exist, we take the average of the probabilities for reaching them.  

Results are presented in Figure~\ref{fig:boxplots} as boxplots over 1,000 reduced fitness landscapes for each perturbation size N (which is also the size of the restricted problem) and NoNSF, NSF properties. For easy comparison, medians are presented separately in Figure~\ref{tab:medP}. 

When N=3, we observe that perturbation is on average ineffective for both types of landscapes, since the global optimum of the restricted problem is reached with a median probability of 1. This means that such a small perturbation does not allow the search algorithm to find new basins of attraction that lead to other local optima of the complete problem for both NSF and NoNSF. It is worth noting, however, that while in all NSF landscapes the search could not escape the global optimum of the restricted problem, in some of the NoNSF landscapes the perturbation was effective. 

For NoNSF landscapes the probability of reaching a global optimum of the restricted problem drops from 1.0 for a chosen set of 3 variables to 0.01 for 9 variables. This means that, for landscapes that do not satisfy the NSF property, increasing the perturbation size, increases the probability of escaping a local optimum. This contrasts with NSF landscapes where even with a 9-variable chosen set, all NSF landscape have a probability above 0.9. It is clear that perturbation is not as effective in NSF landscapes compared to NoNSF landscapes.

To check for a statistical difference of the results from NoNSF and NSF landscapes, we use the Kolmogorov-Smirnov (KS) nonparametric test~\cite{Pettitt77S}. The KS test is used since the distributions are not normal. 
We submitted the results from the 1,000 fitness landscapes from the two different types (NSF and NoNSF) to the KS analysis with a null hypothesis of no difference between the results. All KS tests used for establishing differences under the assumption that the results are not normally distributed, result in a rejection of the null hypothesis at a 99\% confidence level. Hence we conclude that the difference in the results is statistically significant, and the effectiveness of perturbation is impacted by the NSF property.

The results make it clear that it is much harder to escape from a local optimum when the landscape has NSF. The probability of escaping from local optima depends on the structure of the basins of attraction. The NSF property forces solutions with similar fitness to be neighbours, and as a results impacts the structure of the basins, which in turn, impacts the probability of escaping from local optima. Even for N=9, which allows for large perturbations, the probability of escaping such a local optimum is not higher than 1\%. This indicates that, in order to escape a local optimum in NSF landscapes, perturbation is not an effective strategy.

\section{Conclusion}
\label{conclusion}

Search is at the core of many AI problems, such as learning, SAT and optimisation. Search methods are prone to premature convergence, leading to suboptimal solutions, hence most search methods incorporate in their design restart strategies such as perturbation. This research provides an analysis of the conditions under which perturbation is effective, thus helping algorithm designers and practitioners choose the right restart strategy.  

The experimental evaluation on 14,000 fitness landscapes show that in typical landscapes that have the NSF property, once a local optimum has been found, it is necessary to restart the search \emph{further} from the previous local optimum than in a landscape without NSF. 

Even perturbing 9 variables, only 10 out of the 1,000 randomly created fitness landscapes with NSF could escape from a local optimum. On the other hand, our analysis shows that it is much easier to escape from local optima in NoNSF landscapes. It is possible that the tighter the NSF property (i.e. the more correlated the fitness of neighbours in a landscape) the harder it is to escape from local optima. We plan to explore this hypothesis in future work.

Our model of a problem and its landscape is abstract. Solutions of the same fitness are not distinguished, and the only information about the landscape is the number of neighbours of any given fitness that have each other fitness value. This level of abstraction frees us from concerns about specific local search algorithms, and enables us to address general mathematical properties.

\section*{References}
\bibliographystyle{elsarticle-num}
\bibliography{references,lion2017,greedymaximization2018aaai} 

\begin{thebibliography}{10}
\expandafter\ifx\csname url\endcsname\relax
  \def\url#1{\texttt{#1}}\fi
\expandafter\ifx\csname urlprefix\endcsname\relax\def\urlprefix{URL }\fi
\expandafter\ifx\csname href\endcsname\relax
  \def\href#1#2{#2} \def\path#1{#1}\fi

\bibitem{SCHUURMANS2001121}
D.~Schuurmans, F.~Southey, Local search characteristics of incomplete sat
  procedures, Artificial Intelligence 132~(2) (2001) 121 -- 150.

\bibitem{ZIVAN20141}
R.~Zivan, S.~Okamoto, H.~Peled, Explorative anytime local search for
  distributed constraint optimization, Artificial Intelligence 212 (2014) 1 --
  26.

\bibitem{moser2016investigating}
I.~Moser, M.~Gheorghita, A.~Aleti, Investigating the correlation between
  indicators of predictive diagnostic optimisation and search result quality,
  Information Sciences 372 (2016) 162--180.

\bibitem{moser2016identifying}
I.~Moser, M.~Gheorghita, A.~Aleti, Identifying features of fitness landscapes
  and relating them to problem difficulty, Evolutionary Computation 25~(3)
  (2017) 407--437.

\bibitem{lourencco2019iterated}
H.~R. Louren{\c{c}}o, O.~C. Martin, T.~St{\"u}tzle, Iterated local search:
  Framework and applications, in: Handbook of metaheuristics, Springer, 2019,
  pp. 129--168.

\bibitem{Weinberger90}
E.~D. Weinberger, Correlated and uncorrelated fitness landscapes and how to
  tell the difference, Biological Cybernetics 63 (1990) 325--336.

\bibitem{Weinberger91b}
E.~D. Weinberger, Local properties of {Kauffman's} {N}-k model: {A} tunably
  rugged energy landscape, Physical Review A 44~(10) (1991) 6399--6413.

\bibitem{Angel98}
E.~Angel, V.~Zissimopoulos, Autocorrelation coefficient for the graph
  bipartitioning problem, Theoretical Computer Science 191 (1998) 229--243.

\bibitem{Vassilev00}
V.~Vassilev, T.~Fogarty, J.~Miller, Information characteristics and the
  structure of landscapes, Evolutionary Computation 8(1) (2000) 31--60.

\bibitem{Jones95a}
T.~Jones, S.~Forrest, Fitness distance correlation as a measure of problem
  difficulty for genetic algorithms, in: International Conference on Genetic
  Algorithms, 1995, pp. 184--192.

\bibitem{Merz00}
P.~Merz, B.~Freisleben, Fitness landscapes, memetic algorithms and greedy
  operators for graph bipartitioning, Evolutionary Computation 8 (2000) 61--91.

\bibitem{Czech08}
Z.~Czech, Statistical measures of a fitness landscape for the vehicle routing
  problem, in: IEEE International Symposium on Parallel and Distributed
  Processing, 2008, pp. 1--8.

\bibitem{Merz04}
P.~Merz, Advanced fitness landscape analysis and the performance of memetic
  algorithms, Evolutionary Computation 12~(3) (2004) 303--325.

\bibitem{Altenberg97}
L.~Altenberg, Fitness distance correlation analysis: An instructive
  counterexample, in: International Conference on Genetic Algorithms, Morgan
  Kaufmann, 1997, pp. 57--64.

\bibitem{Quick98}
R.~Quick, J.~Rayward-Smith, G.~Smith, Fitness distance correlation and ridge
  functions, in: Parallel Problem Solving from Nature, Vol. 1498, Springer,
  1998, pp. 77--86.

\bibitem{Naudts00}
B.~Naudts, L.~Kallel, A comparison of predictive measures of problem difficulty
  in evolutionary algorithms, IEEE Transactions on Evolutionary Computation
  4~(1) (2000) 1--15.

\bibitem{daolio2010local}
F.~Daolio, S.~Verel, G.~Ochoa, M.~Tomassini, Local optima networks of the
  quadratic assignment problem, in: Congress on Evolutionary Computation, IEEE,
  2010, pp. 1--8.

\bibitem{Mar:bc:03}
R.~Marti, Multi-start methods, in: F.~Glover, G.~A. Kochenberger (Eds.),
  Handbook of Metaheuristics, 2003, pp. 355--368.

\bibitem{Lou-Mar-Stu:bc:10}
H.~R. Lourenço, O.~C. Martin, T.~St{\"u}tzle, Iterated local search: Framework
  and applications, in: Handbook of Metaheuristics, Springer, 2010, pp.
  363--397.

\bibitem{GomesSCK00}
C.~P. Gomes, B.~Selman, N.~Crato, H.~A. Kautz, Heavy-tailed phenomena in
  satisfiability and constraint satisfaction problems, Journal of Automated
  Reasoning 24~(1) (2000) 67--100.

\bibitem{luby1993}
M.~Luby, A.~Sinclair, D.~Zuckerman, Optimal speedup of {Las Vegas} algorithms,
  Information Processing Letters 47~(4) (1993) 173--180.

\bibitem{jansen2002dynrestartsEA}
T.~Jansen, On the analysis of dynamic restart strategies for evolutionary
  algorithms, in: Parallel Problem Solving from Nature, Springer Berlin
  Heidelberg, 2002, pp. 33--43.

\bibitem{Sch-Tey-Tey:c:12}
M.~Schoenauer, F.~Teytaud, O.~Teytaud, A rigorous runtime analysis for
  quasi-random restarts and decreasing stepsize, in: Artificial Evolution,
  Springer, 2012, pp. 37--48.

\bibitem{Biere08}
A.~Biere, Adaptive restart strategies for conflict driven {SAT} solvers, in:
  Theory and Applications of Satisfiability Testing, 2008, pp. 28--33.

\bibitem{Huang2007}
J.~Huang, The effect of restarts on the efficiency of clause learning, in:
  International Joint Conference on Artificial Intelligence, 2007, pp.
  2318--2323.

\bibitem{CireKS14}
A.~A. Cir{\'{e}}, S.~Kadioglu, M.~Sellmann, Parallel restarted search, in:
  {AAAI} Conference on Artificial Intelligence, 2014, pp. 842--848.

\bibitem{Ansotegui2017dialectic}
C.~Ansótegui, J.~Pon, M.~Sellmann, K.~Tierney, Reactive dialectic search
  portfolios for {MaxSAT}, in: AAAI Conference on Artificial Intelligence,
  2017, pp. 765--772.

\bibitem{Neumann2010eaTheoryBook}
F.~Neumann, C.~Witt, Bioinspired Computation in Combinatorial Optimization,
  Natural Computing Series, Springer, 2010.

\bibitem{Doerr2017fastGAs}
B.~Doerr, H.~P. Le, R.~Makhmara, T.~D. Nguyen, Fast genetic algorithms, in:
  Genetic and Evolutionary Computation Conference, ACM, 2017, pp. 777--784.

\bibitem{Friedrich2018heavytailedPPSN}
T.~Friedrich, A.~G{\"o}bel, F.~Quinzan, M.~Wagner, Heavy-tailed mutation
  operators in single-objective combinatorial optimization, in: Parallel
  Problem Solving from Nature, Springer, 2018, pp. 134--145.

\bibitem{Friedrich2018heavytailedGECCO}
T.~Friedrich, F.~Quinzan, M.~Wagner, Escaping large deceptive basins of
  attraction with heavy-tailed mutation operators, in: Genetic and Evolutionary
  Computation Conference, ACM, 2018, pp. 293--300.

\bibitem{trivedi1982probability}
K.~S. Trivedi, Probability and statistics with reliability, queuing, and
  computer science applications, Vol.~13, Wiley Online Library, 1982.

\bibitem{sinclair1993markov}
A.~Sinclair, Markov chains and rapid mixing, in: Algorithms for Random
  Generation and Counting: A Markov Chain Approach, Springer, 1993, pp. 42--62.

\bibitem{weise2019improved}
T.~Weise, Z.~Wu, M.~Wagner, An improved generic bet-and-run strategy with
  performance prediction for stochastic local search, in: AAAI Conference on
  Artificial Intelligence, 2019, pp. 2395--2402.

\bibitem{jeavons1995tractable}
P.~G. Jeavons, M.~C. Cooper, Tractable constraints on ordered domains,
  Artificial Intelligence 79~(2) (1995) 327--339.

\bibitem{Pettitt77S}
A.~N. Pettitt, M.~A. Stephens, The {Kolmogorov-Smirnov} goodness-of-fit
  statistic with discrete and grouped data, Technometrics 19~(2) (1977)
  205--210.

\end{thebibliography}
\end{document}